\documentclass[runningheads]{llncs}
\usepackage{graphicx}
\usepackage{cite} 
\usepackage{times}
\usepackage{epsfig}
\usepackage{amsmath}
\usepackage{amssymb}
\usepackage{mwe}
\usepackage{acro}
\usepackage{amssymb}
\usepackage{xcolor,colortbl}
\usepackage{tabularx}
\usepackage{relsize}
\usepackage{pifont}
\usepackage{booktabs} 
\usepackage{multirow}
\usepackage{multicol}
\usepackage{adjustbox}
\usepackage{float}
\usepackage[colorlinks,
            linkcolor=red,  
            anchorcolor=blue, 
            citecolor=green,   
            ]{hyperref}

\DeclareAcronym{ROI}{
short=ROI,
long=region of interest,
}

\DeclareAcronym{IOU}{
short=IOU,
long=intersection over union,
}

\DeclareAcronym{cIOU}{
short=cIOU,
long=circle intersection over union,
}

\DeclareAcronym{DoF}{
short=DoF,
long=degrees of freedom,
}

\begin{document}
\title{Compound Figure Separation of Biomedical Images with Side Loss}
%
\author{Tianyuan Yao\inst{1} \and
Chang Qu \inst{1} \and 
Quan Liu \inst{1} \and
Ruining Deng \inst{1} \and
Yuanhan Tian \inst{1} \and
Jiachen Xu \inst{1} \and
Aadarsh Jha \inst{1} \and
Shunxing Bao \inst{1} \and
Mengyang Zhao \inst{3} \and
Agnes B. Fogo \inst{2} \and
Bennett A. Landman \inst{1} \and
Catie Chang \inst{1} \and
Haichun Yang \inst{2} \and
Yuankai Huo\inst{1}}

\authorrunning{H. Yang et al.}
\institute{Vanderbilt University, Nashville TN 37215, USA \and
Vanderbilt University Medical Center, Nashville TN 37215, USA \and
Tufts University, Medford MA 02155, USA}
\maketitle              
\begin{abstract}
Unsupervised learning algorithms (e.g., self-supervised learning, auto-encoder, contrastive learning) allow deep learning models to learn effective image representations from large-scale unlabeled data. In medical image analysis, even unannotated data can be difficult to obtain for individual labs. Fortunately, national-level efforts have been made to provide efficient access to obtain biomedical image data from previous scientific publications. For instance, NIH has launched the Open-i$^\circledR$ search engine that provides a large-scale image database with free access. However, the images in scientific publications consist of a considerable amount of compound figures with subplots. To extract and curate individual subplots, many different compound figure separation approaches have been developed, especially with the recent advances in deep learning. However, previous approaches typically required resource extensive bounding box annotation to train detection models. In this paper, we propose a simple compound figure separation (SimCFS) framework that uses weak classification annotations from individual images. Our technical contribution is three-fold: (1) we introduce a new side loss that is designed for compound figure separation; (2) we introduce an intra-class image augmentation method to simulate hard cases; (3) the proposed framework enables an efficient deployment to new classes of images, without requiring resource extensive bounding box annotations. From the results, the SimCFS achieved a new state-of-the-art performance on the ImageCLEF 2016 Compound Figure Separation Database. The source code of SimCFS is made publicly available at https://github.com/hrlblab/ImageSeperation.
\end{abstract}
\section{Introduction}

\begin{figure}[t]
\begin{center}
\includegraphics[width=0.75\linewidth]{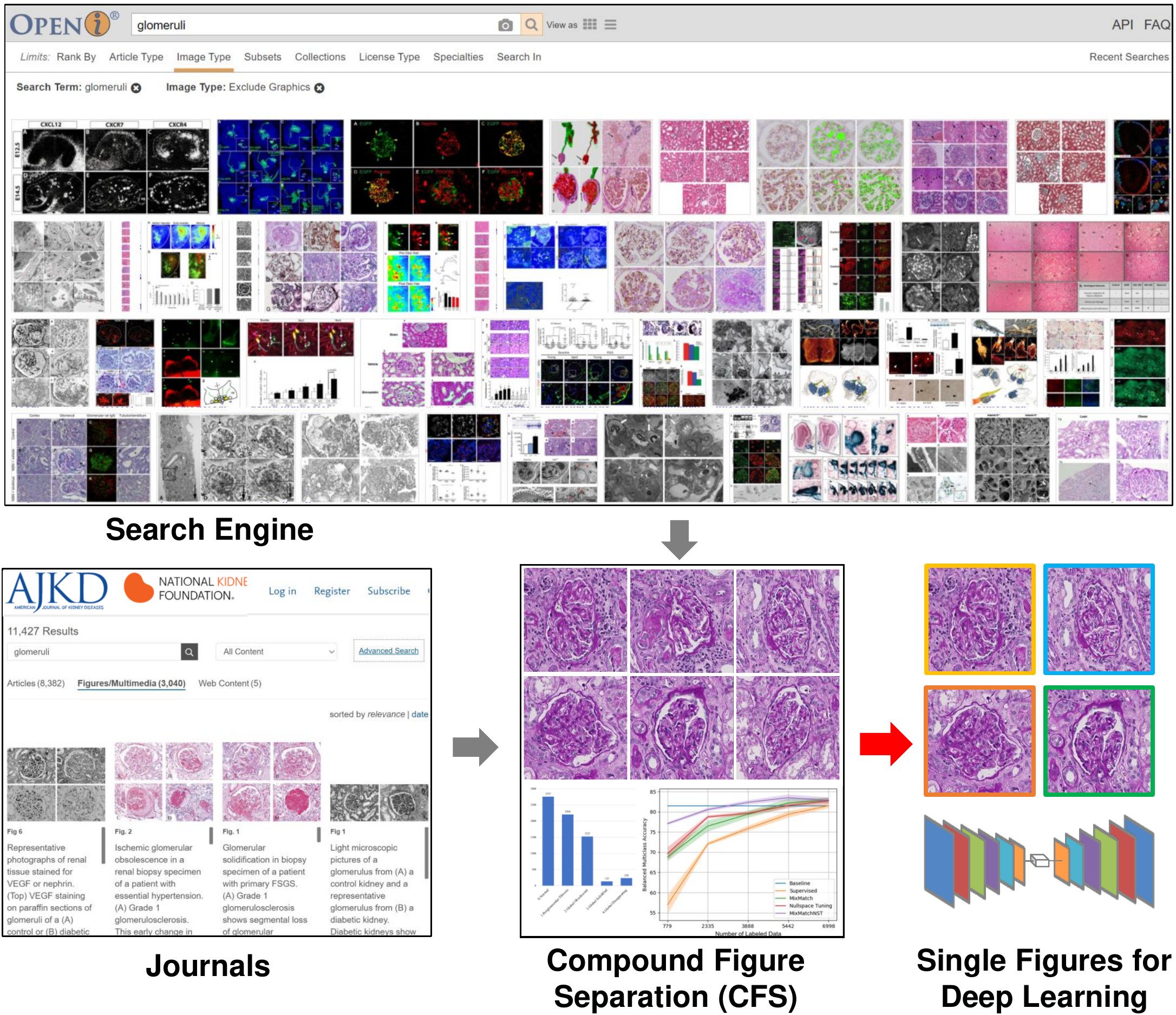}
\end{center}
   \caption{\textbf{Value of compound figure separation.} This figure shows the hurdle (red arrow) of training self-supervised machine learning algorithms directly using large-scale biomedical image data from biomedical image databases (e.g., NIH OpenI) and academic journals (e.g., AJKD). When searching desired tissues  (e.g., search ``glomeruli"), a large amount of data are compound figures. Such data would advance medical image research via recent unsupervised learning algorithms, such as self-supervised learning, contrasting learning, and auto encoder networks~\cite{huo2021ai}}
\label{fig:idea}
\end{figure}

Unsupervised learning algorithms\cite{celebi2016unsupervised}\cite{sathya2013comparison} allow deep learning models to learn effective image representations from large-scale unlabeled data , such as self-supervised learning, auto-encoder, and contrastive learning\cite{chen2020simple}. However, even large-scale unannotated glomerular images can be difficult to obtain for individual labs\cite{zhang2017deep}. Fortunately, many resources (e.g., NIH  Open-i$^\circledR$\cite{demner2012design} 
search engine, academic images released by journals) have provided the opportunity to obtain extra large-scale images. However, the images from such resources consist of a considerably large amount of compound figures with subplots (Fig.\ref{fig:idea}). To extract and curate individual subplots, compound figure separation algorithms can be applied\cite{lee2015dismantling}.

Various compound figure separation approaches have been developed\cite{davila2020chart,lee2015detecting,apostolova2013image,tsutsui2017data,shi2019layout,jiang2021two,huang2005associating}, especially with recent advances in deep learning. However, previous approaches typically required resource extensive bounding box annotation to train detection models. In this paper, we propose a simple compound figure separation (SimCFS) framework that utilizes weak classification annotations from individual images for compound figure separation. Briefly, the contribution of this study are in three-fold:

$\bullet$ We propose a new side loss, an optimized detection loss for figure separation.

$\bullet$ We introduce an intra-class image augmentation method to simulate hard cases.

$\bullet$ The proposed framework enables an efficient deployment to new classes of images, without requiring resource extensive bounding box annotations.

We apply our technique to conduct compound figure separation for renal pathology. Glomerular phenotyping\cite{koziell2002genotype} is a fundamental task for efficient diagnosis and quantitative evaluations in renal pathology. Recently, deep learning techniques have played increasingly important roles in renal pathology to reduce the clinical working load of pathologists and enable large-scale population based research~\cite{gadermayr2017cnn,bueno2020glomerulosclerosis,govind2018glomerular,kannan2019segmentation,ginley2019computational}. Due to the lack of publicly available annotated dataset for renal pathology, the related deep learning approaches are still limited on a small-scale~\cite{huo2021ai}. Therefore, it is appealing to extract large-scale glomerular images from public databases (e.g., NIH Open-i$^\circledR$ search engine) for downstream unsupervised or semi-supervised learning~\cite{huo2021ai}.

\section{Related Work}
In biomedical articles, about 40-60$\%$ of figures are multi-panel~\cite{kalpathy2015evaluating}. Several works have been proposed in the document analysis community, extracting figure and semantic information. For example, Huang et al.~\cite{huang2005associating} presented their recognition results of textual and graphical information in literary figures. Davila et al.~\cite{davila2020chart} presented a survey of approaches of several data mining pipelines for future research.

In order to collect scientific data massively and automatically, various approaches have been proposed by different researchers\cite{10.1093/bioinformatics/btx611} \cite{10.1007/978-3-319-65813-1_20}\cite{lee2015dismantling}. For example, Lee. et al. (2015)~\cite{lee2015detecting} proposed an SVM-based binary classifier to distinguish complete charts from visual markers like labels, legend, and ticks. Apostolova et al.~\cite{apostolova2013image} proposed a figure separation method by capital index. These traditional computer vision approaches are commonly based on the figure's grid-based layout or visual information. Thus, the separation was usually accomplished by an x-y cut. However there are more complicated cases in compound figures like no white-space gaps or overlapped situations. 

In the past few years, recent deep learning based algorithms using convolutional neural networks(CNNs) provided considerably better performance in extracting and processing textual and non-textual content from scholarly articles. Tsutsui and Crandall (2017)~\cite{tsutsui2017data} proposed the first deep learning based approach to compound figure separation in which they applied a deep convolutional network to train the separator. They also implemented training on artificially-constructed datasets and reported superior performances on ImageCLEF data sets\cite{GSB2016}. Shi et al.~\cite{shi2019layout} developed a multi-branch output CNN to predict the irregular panel layouts and provided augmented data to drive learning; their network can predict compound figures of different sizes of structures with a better accuracy. 
Jiang et al.~\cite{jiang2021two} combined the traditional vision method and high performance of deep learning networks by firstly detecting the sub-figure label and then optimizing the feature selection process in the sub-figure detection part. This improved the detection precision by 9$\%$.In Tsutsui's study~\cite{tsutsui2017data}, they applied You Only Look Once (YOLO) Version 2~\cite{redmon2016you}, a CNN based detection network. Deep learning based detection approaches utilized a single convolutional network to predict bounding boxes and class probabilities from full images simultaneously, which can achieve high speed detection and are in favor of sub-figure detection tasks.
\begin{figure}[t]
\begin{center}
\includegraphics[width=0.8\linewidth]{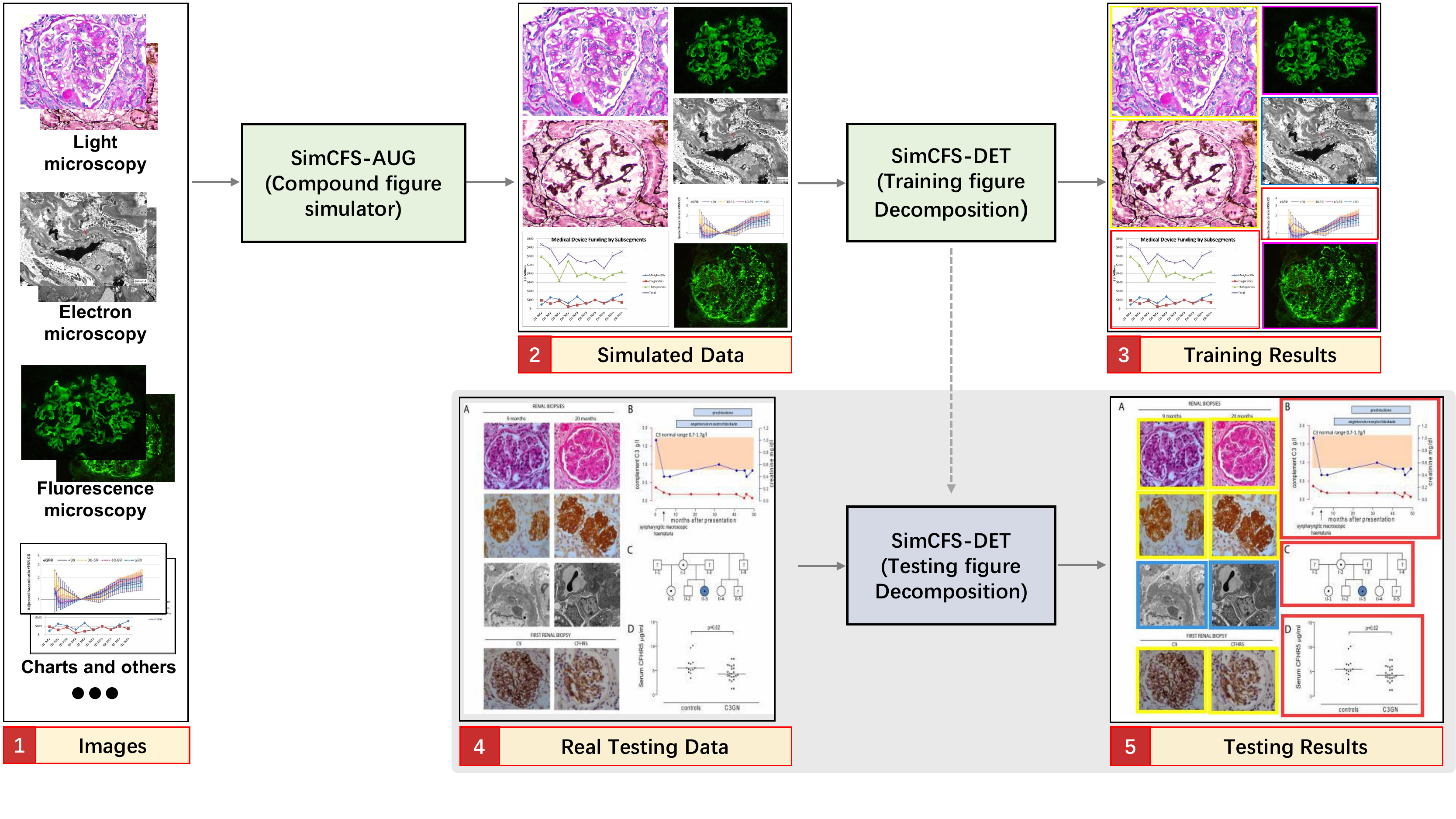}
\end{center}
   \caption{\textbf{The overall workflow of the proposed simple compound figure separation (SimCFS) workflow.} In the training stage, SimCFS only requires single images from different categories. The pseudo compound figures are generated from the proposed augmentation simulator (SimCFS-AUG). Then, a detection network (SimCFS-DET) is trained to perform compound figure separation. In the testing stage (the gray panel), only the trained SimCFS-DET is required for separating the images.}
\label{fig:network}
\end{figure}

\section{Methods}
The overall framework of the SimCFS approach is presented in Fig.~\ref{fig:network}. The training stage of SimCFS contains two major steps: (1) compound figure simulation, and (2) sub-figure detection. In the testing stage, only the detection network is needed.

\subsection{Anchor based detection}
YOLOv5, the latest version in the YOLO family~\cite{bochkovskiy2020yolov4}, is employed as the backbone network for sub-figure detection. The rationale for choosing YOLOv5 is that the sub-figures in compound figures are typically located in horizontal or vertical orders. Herein, the grid-based design with anchor boxes is well adaptable to our application. A new side loss is introduced to the detection network that further optimizes the performance of compound figure separation. 

\subsection{Compound figure simulation}
Our goal is to only utilize single images, which are non-compound images with weak classification labels in training a compound image separation method. In previous studies, the same task typically requires stronger bounding box annotations of subplots using real compound figures. In compound figure separation tasks, a unique advantage is that the sub-figures are not overlapped. Moreover, their spatial distributions are more ordered compared with natural images in object detection. Therefore, we propose to directly simulate compound figures from individual images as the training data for the downstream sub-figure detection.

\begin{figure}[t]
\begin{center}
\includegraphics[width=1\linewidth]{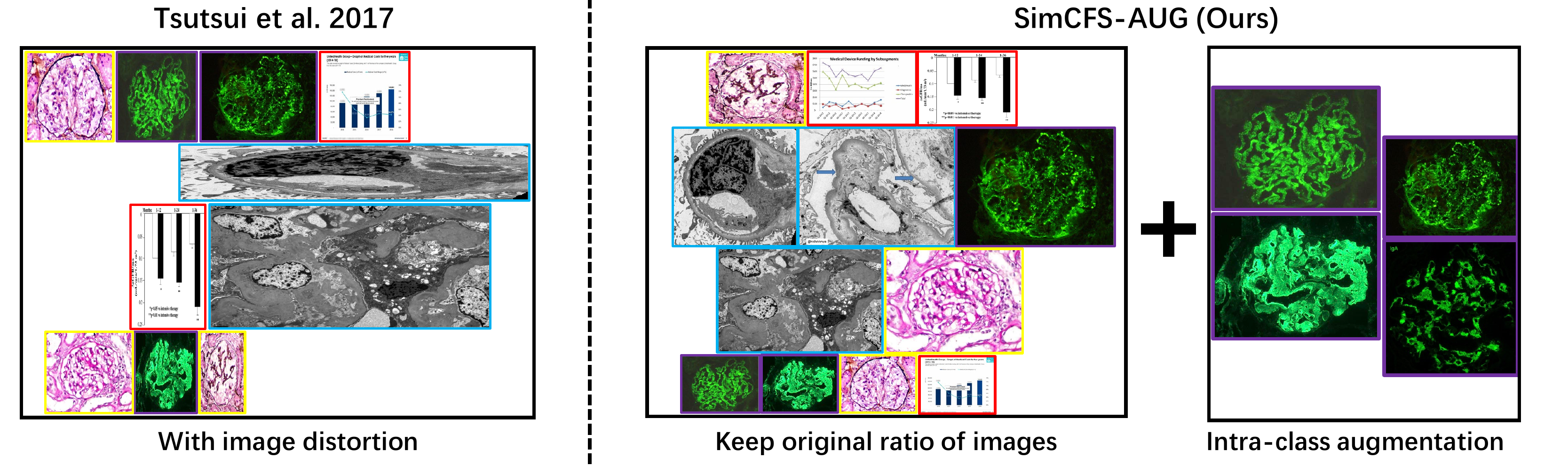}
\end{center}
   \caption{\textbf{Compound figure simulation.} The left panel shows the current compound figure simulation strategy, which distorts the images with random ratios. The right panel presents the proposed SimCFS-AUG compound figure simulator. It keeps the original ratio of individual images. Moreover, intra-class augmentation is introduced to simulate the hard cases that the figures with similar appearances attach to each other.}
\label{fig:method}
\end{figure}

Tsutsui et al.~\cite{tsutsui2017data} proposed a compound figure synthesis approach (Fig.~\ref{fig:method}). The method first randomly samples a number of rows and random height for each row. Then a random number of single figures fills the empty template. However, the single figures are naively resized to fit the template, with large distortion (Fig.~\ref{fig:method}).

Inspired by prior arts~\cite{tsutsui2017data}, we propose a simple compound figure separation specific data augmentation strategy, called SimCFS-AUG, to perform compound figure simulation. Two groups of simulating compound figures are generated which are row-restricted and column-restricted. The length of each row or column is randomly generated within a certain range. Then, images from our database are randomly selected and concatenated together to fit in the preset space. As opposed to previous studies, the original ratio of individual images is kept in our SimCFS-AUG simulator, without distortion. Moreover, we introduce a new class within compound image separation augmentation to SimCFS-AUG so as to simulate the specific hard case in which all images belong to the same class.

\begin{figure}[t]
\begin{center}
 \includegraphics[width=0.8\linewidth]{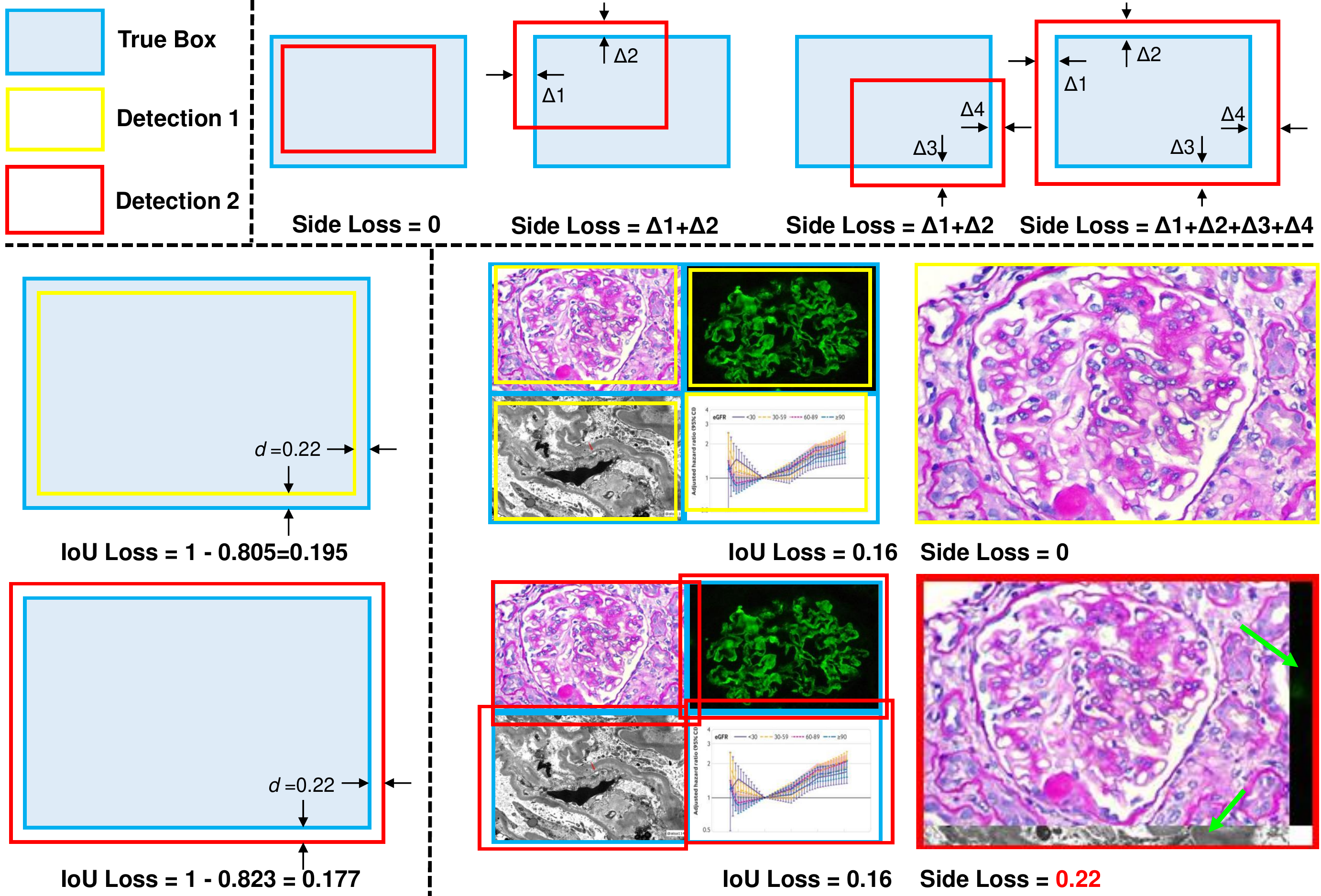}
\end{center}
   \caption{\textbf{Proposed Side loss for figure separation}. The upper panel shows the principle of side loss, in which penalties only apply when vertices of detected bounding boxes are outside of true box regions. The lower left panel shows the bias of current IoU loss towards the over detection. When an under detection case (yellow box) and an over detection case (red box) have the same margins ($d$), from predicted to true boxes, the over detection has the smaller loss (larger IoU). The lower right panel shows that the under detection and over detection examples of the compound figure separation, with the same IoU loss. The Side loss is proposed to break the even IoU loss, given the results in the yellow boxes are less contaminated by nearby figures than the results in the red boxes (green arrows).}
\label{fig:sideloss}
 \end{figure}

\subsection{Side loss for compound figure separation}
For object detection on natural images, there is no specific preference between over detection and under detection as objects can be randomly located and even overlapped. In medical compound images, however, the objects are typically closely attached to each other, but not overlapping. In this case, over detection would introduce undesired pixels from the nearby plots (Fig. ~\ref{fig:sideloss}), which are not ideal for downstream deep learning tasks. Unfortunately, the over detection is often encouraged by the current Intersection Over Union (IoU) loss in object detection (Fig. ~\ref{fig:sideloss}), compared with under detection.

In the SimCFS-DET network, we introduce a simple side loss, which will penalize over detection. We define a predicted bounding box as $B^p$ and a ground truth box as $B^g$, with coordinates: $B^p = (x^p_1,y^p_1,x^p_2,y^p_2) $,$\quad B^g = (x^g_1,y^g_1,x^g_2,y^g_2)$. The over detection penalty of vertices for each box is computed as:
\begin{equation}
\begin{aligned}
x^{\mathcal{I}}_1 = \max(0,x^g_1-x^p_1),
y^{\mathcal{I}}_1 = \max(0,y^g_1-y^p_1)\\
x^{\mathcal{I}}_2 = \max(0,x^p_2-x^g_2),
y^{\mathcal{I}}_2 = \max(0,y^p_2-y^g_2)
\end{aligned}
\end{equation}

Then, the side loss is defined as:
\begin{equation}
\mathcal{L}_{side} = x^{\mathcal{I}}_1+y^{\mathcal{I}}_1+x^{\mathcal{I}}_2+y^{\mathcal{I}}_2
\end{equation}

Side loss is combined with canonical loss functions in YOLOv5, including bounding box loss (${L}_{box}$), object probability loss (${L}_{obj}$), and classification loss (${L}_{cls}$).\\ 
{$ \mathcal{L}_{total} = \lambda_1{L}_{box} + \lambda_2{L}_{obj} + \lambda_3{L}_{cls} + \lambda_4{L}_{side}
$}   ,where $\lambda_1$, $\lambda_2$, $\lambda_3$, $\lambda_4$ are constant weights to balance the four loss functions. Following the YOLOv5's implementation \footnote{https://github.com/ultralytics/yolov5}, the parameters were set as $\lambda_1$ = ${box}\times(3/{nl})$, $\lambda_2$ = ${obj}\times{({imgsize}/640)^{2}\times(3/{nl})}$, $\lambda_3$ = $({cls}\times{num\_cls}/80)\times(3/{nl})$, where ${num\_cls}$ was the number of classes, ${nl}$ was the number of layers, and ${imgsize}$ was the image size.The $\lambda_4$ of the Side loss was empirically set to $\lambda_1/30$ across all experiments as the Side loss and Box loss are all based on the coordinates.

\begin{figure}[t]
\begin{center}
\includegraphics[width=0.9\linewidth]{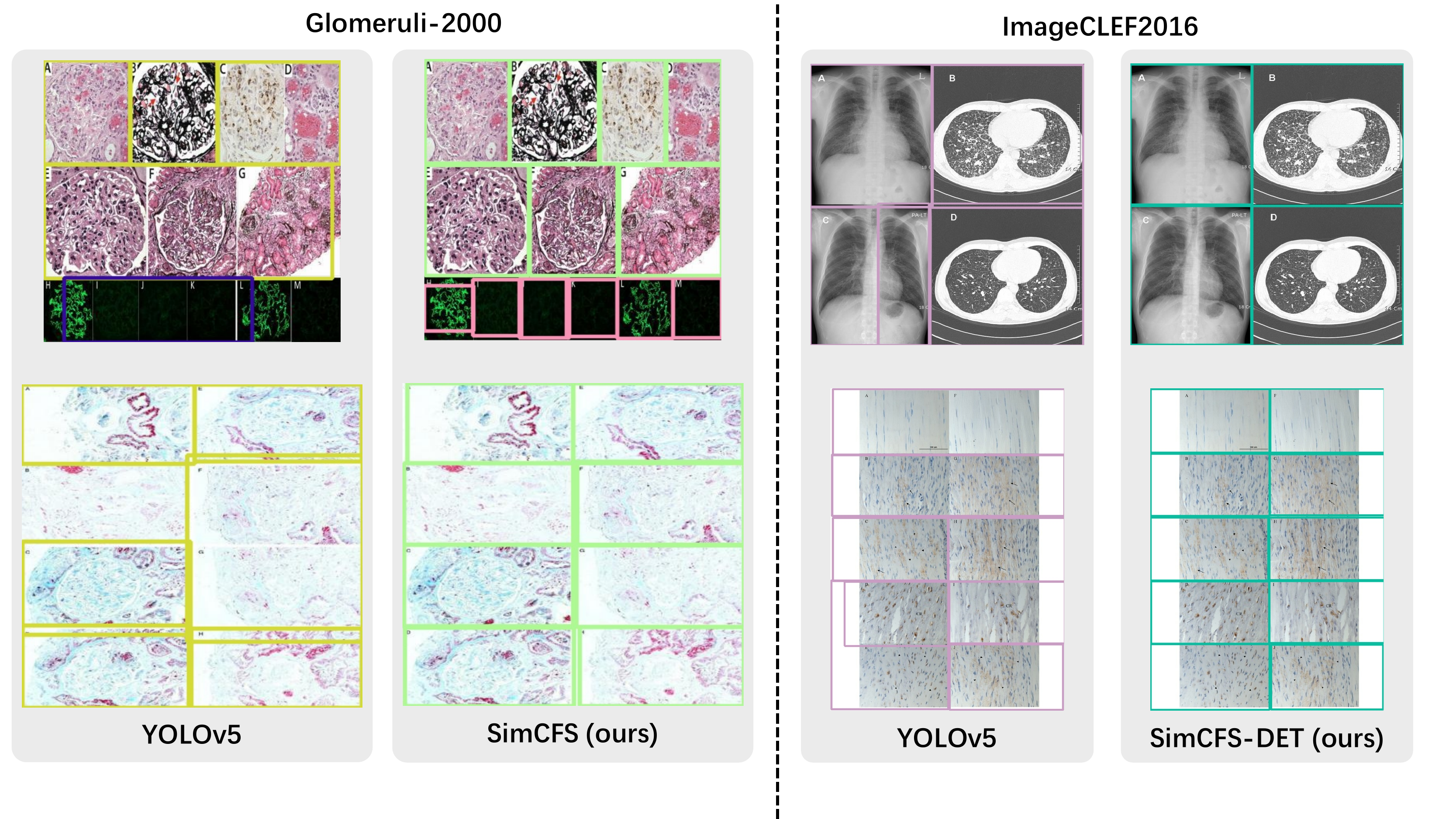}
\end{center}
   \caption{\textbf{Qualitative Results.} This figure shows the qualitative results of comparing proposed SimCFS approach with the YOLOv5 benchmark.}
\label{fig:results}
\end{figure}

\section{Data and Implementation Details}
We collected two in-house datasets for evaluating the performance of different compound figure separation strategies. One compound figure dataset (called Glomeruli-2000) consisted of 917 training and 917 testing real figure plots from the American Journal of Kidney Diseases (AJKD), with the keywords ``glomerular OR glomeruli OR glomerulus" as the keywords. Each figure was annotated manually with four classes, including glomeruli from (1) light microscopy, (2) fluorescence microscopy,  and (3) electron microscopy, and (4) charts/plots. 

To obtain single images to simulate compound figures, we downloaded 5,663 single individual images from other resources. Briefly, we obtained 1,037 images from Twitter, and obtained 4,626 images from the Google search engine, with five classes, including single images from (1) glomeruli with light microscopy, (2) glomeruli with fluorescence microscopy, (3) glomeruli with electron microscopy, (4) charts/plots, and (5) others. The individual images were combined using the SimCFS-AUG simulator to generate 9,947 pseudo training images. 2,000 of the pseudo images were simulated using intra-class augmentation, while 2,947 of them were simulated with only single sub-figures. The implementation of SimCFS-DET was based on YOLOv5 with PyTorch implementations. Google Colab was used to perform all experiments in this study.

In the experiment setting, the parameters are empirically chosen. We set the learning rate to 0.01, weight decay to 0.0005 and momentum to 0.937. The input image size was set to 640, ${box}$ to 0.5, ${obj}$ to 1, ${cls}$ to 0.5, and number of layers to 3. For our in-house datasets, we trained 50 epochs using a batch size of 64. For the imageCLEF2016 dataset\cite{GSB2016}, we trained 50 epochs using a smaller batch size of 8.

\section{Results}
\subsection{Ablation Study}
The Side loss is the major contribution to the YOLOv5 detection backbone. In this ablation study, we show the performance of using 917 real compound images with manual box annotations as training data (as ``Real Training Images'') in Table \ref{table:ablation} and Fig.~\ref{fig:results}. This also shows the results of merely using simulated images as training data (as ``Simulated Training Images''). The proposed side loss consistently improves the detection performance by a decent margin. The intra-class self-augmentation improves the performance when only using simulated training images.

\begin{table}[t]
\caption{The ablation study with different types of training data.}
\begin{tabular}{l@{}c@{\ \ }p{1.3cm}p{1cm}p{0.8cm}p{0.8cm}p{0.8cm}p{0.8cm}p{0.8cm}}
\toprule
Training Data& Method &Side loss & AUG & All  & Light & Fluo.  & Elec. &Chart \\
\midrule
\multirow{2}{0.8in}{Real Training Images} 
&YOLOv5~\cite{bochkovskiy2020yolov4}   &   &       & 69.8      & 77.1 & 71.3       & 73.4 & 57.4\\
&SimCFS-DET (ours)    & \checkmark &      & 79.2  & 86.1      & \textbf{80.9}     & 84.2 & \textbf{65.8}  \\
\midrule
\multirow{4}{0.8in}{Simulated Training Images}
&YOLOv5~\cite{bochkovskiy2020yolov4}      &   &   & 66.4      & 79.3 & 62.1       & 76.1    & 48.0\\
&SimCFS (ours)      & \checkmark &    & 69.4      & 77.6 & 67.1       & 84.1  & 48.8 \\
&YOLOv5~\cite{bochkovskiy2020yolov4} &   &    \checkmark & 71.4      & 82.8 & 72.1       & 75.3  &47.1   \\
&SimCFS (ours) & \checkmark  & \checkmark     & \textbf{80.3}      & \textbf{89.9} & 78.7       & \textbf{87.4}  &58.8\\
\bottomrule
\end{tabular}

*AUG is the intra-class self-augmentation. ALL is the Overall mAP$_{0.5:.95}$, which is reported for all classes, class Light, class Florescence and class Electron.
\label{table:ablation}
\end{table}

\begin{table}[t]
\centering
\caption{The results on ImageCLEF2016 dataset.}
\begin{tabular}{p{3cm}p{2cm}p{1.2cm}p{1.2cm}p{1.2cm}p{1.2cm}}
\toprule
Method      & Backbone          & mAP$_{0.5}$ & mAP$_{0.5:.95}$ \\
\midrule
Tsutsui et al.~\cite{tsutsui2017data}   & YOLOv2      & 69.8 & -       \\
Tsutsui et al.~\cite{tsutsui2017data}  & Transfer  & 77.3 & -       \\
Zou et al.~\cite{zou2020unified}   & ResNet152      & 78.4 & -       \\
Zou et al.~\cite{zou2020unified}  & VGG19          & 81.1 & -       \\
YOLOv5~\cite{bochkovskiy2020yolov4}  & YOLOv5   & 85.3 & 69.5   \\
SimCFS-DET (ours)   &YOLOv5  & \textbf{88.9} & \textbf{71.2}  \\
\bottomrule
\end{tabular}
\label{table:stateoftheart}
\end{table}

\subsection{Comparison with State-of-the-art}
We also compare CFS-DET with the state-of-the-art approaches including Tsuisui et al.~\cite{tsutsui2017data} and Zou et al.~\cite{zou2020unified} using the ImageCLEF2016 dataset\cite{GSB2016}. ImageCLEF2016 is the commonly accepted benchmark for compound figure separation, including total 8,397 annotated multi-panel figures (6,783 figures for training and 1,614 figures for testing). Table \ref{table:stateoftheart} shows the results of the ImageCLEF2016 dataset. The proposed CFS-DET approach consistently outperforms other methods by considering  evaluation metrics. 

\section{Conclusion}

In this paper, we introduce the SimCFS framework to extract images of interests from large-scale compounded figures with merely weak classification labels. The pseudo training data can be built using the proposed SimCFS-AUG simulator. The anchor-based SimCFS-DET detection achieves state-of-the-art performance by introducing a simple Side loss. 




%
%
\bibliographystyle{splncs04}
\bibliography{main}
%




\end{document}